\documentclass[sigconf]{acmart}
\usepackage{amsthm}
\usepackage{algorithm}
\usepackage{algorithmic}
\usepackage{caption}
\usepackage{float}
\usepackage{placeins}
\newlength\myindent
\usepackage[most]{tcolorbox}
\usepackage{multicol}
\AtBeginDocument{%
  }
\copyrightyear{2026}
\acmYear{2026}
\setcopyright{cc}
\setcctype{by}
\acmConference[GECCO '26]{Genetic and Evolutionary Computation Conference}{July 13--17, 2026}{San Jose, Costa Rica}
\acmBooktitle{Genetic and Evolutionary Computation Conference (GECCO '26), July 13--17, 2026, San Jose, Costa Rica}
\acmDOI{10.1145/3795095.3805080}
\acmISBN{979-8-4007-2487-9/2026/07}

\begin{document}

\title{Evolution With Purpose: \\Hierarchy-Informed Optimization of Whole-Brain Models}

\author{Hormoz Shahrzad}
\email{hormoz@cs.utexas.edu}
\affiliation{
  \institution{The University of Texas at Austin; Stanford University; Cognizant AI Lab}
  \city{}
  \country{}
}
\author{Niharika Gajawelli}
\email{gajawell@stanford.edu}
\affiliation{
  \institution{Stanford University}
  \city{}
  \country{}
}
\author{Kaitlin Maile}
\authornote{Now at Google}

\email{kmaile@google.com}
\affiliation{
  \institution{University of Toulouse}
  \city{}
  \country{}
}
\author{Manish Saggar}
\email{saggar@stanford.edu}
\affiliation{
  \institution{Stanford University}
  \city{}
  \country{}
}
\author{Risto Miikkulainen}
\email{risto@cs.utexas.edu}
\affiliation{
  \institution{The University of Texas at Austin; Cognizant AI Lab}
  \city{}
  \country{}
}
\renewcommand{\shortauthors}{Shahrzad et al.}
\renewcommand{\shorttitle}{Evolution With Purpose: Hierarchy-Informed Optimization of Whole-Brain Models}

\begin{abstract}
Evolutionary search is well suited for large-scale biophysical brain modeling, where many parameters with nonlinear interactions and no tractable gradients need to be optimized. Standard evolutionary approaches achieve an excellent fit to MRI data; however, among many possible such solutions, it finds ones that overfit to individual subjects and provide limited predictive power. This paper investigates whether guiding evolution with biological knowledge can help. Focusing on whole-brain Dynamic Mean Field (DMF) models, a baseline where 20 parameters were shared across the brain was compared against a heterogeneous formulation where different sets of 20 parameters were used for the seven canonical brain regions. The heterogeneous model was optimized using four strategies: optimizing all parameters at once, a curricular approach following the hierarchy of brain networks (HICO), a reversed curricular approach, and a randomly shuffled curricular approach. While all heterogeneous strategies fit the data well, only curricular approaches generalized to new subjects. Most importantly, only HICO made it possible to use the parameter sets to predict the subjects' behavioral abilities as well. Thus, by guiding evolution with biological knowledge about the hierarchy of brain regions, HICO demonstrated how domain knowledge can be harnessed to serve the purpose of optimization in real-world domains.
\end{abstract}

\ccsdesc[500]{Computing methodologies~Genetic algorithms}
\ccsdesc[500]{Mathematics of computing~Optimization algorithms}
\ccsdesc[300]{Computing methodologies~Bio-inspired approaches}
\ccsdesc[300]{Computing methodologies~Distributed algorithms}
\keywords{%
Evolutionary Computation,
Curriculum Learning,
Dynamic Mean Field Models,
Whole-Brain Modeling,
Cross-Subject Generalization,
Lyapunov Stability,
Human Connectome Project
}


\maketitle

\section{Introduction}
\label{sec:introduction}

Large-scale functional brain dynamics are commonly modeled as emergent phenomena arising from the interaction between anatomical connectivity and local neural physiology. Models such as neural mass, neural field, and Dynamic Mean Field (DMF) successfully reproduced empirical functional connectivity and dynamics at the level of resting-state networks (RSN)s \cite{Honey2009, Deco2011, Breakspear2017, Maile2019}. These models depend on many interacting hyperparameters that must be carefully tuned to match individual neuroimaging data. Evolutionary algorithms are well suited for this purpose because they explore complex, non-differentiable fitness landscapes without gradient information \cite{Hansen2016, Miikkulainen2020, Pelikan2005,Demirtas2019}. Prior work applying evolutionary strategies to DMF models demonstrated improved fits to resting-state data \cite{Maile2019}, but also highlighted that such fits often fail to generalize across subjects and may not predict behavior well. Similar issues of deception and overfitting arise in other real-world search domains. Although evolutionary computation (EC) techniques are well suited for large, high-dimensional, and deceptive search spaces—even at extreme scales involving thousands of states and billions of variables \cite{Deb2017, Gomez1997, Miikkulainen2020, Shahrzad2018, Shahrzad2019}—they can still converge to brittle or uninformative solutions in the absence of appropriate inductive biases.

This work investigates a previously underexplored factor in evolutionary optimization of dynamical systems: the temporal order in which parameters are introduced during search. Drawing on cortical hierarchies, established in humans and primates as a principal gradient of cortical organization \cite{Margulies2016}, and incremental optimization insights \cite{Gomez1997}, heterogeneous optimization was compared with hierarchy-informed curriculum (HICO), reverse-phased curriculum, and shuffled curriculum. Although introducing heterogeneity substantially improved in-sample fitness, only flat (non-curricular) heterogeneous, HICO, and reverse-phased strategies significantly outperformed the homogeneous baseline ($p \approx 10^{-23}$--$10^{-25}$). Marked differences emerged, however, under leave-one-out cross-subject evaluation using cohort-averaged parameters: flat heterogeneous and homogeneous baselines failed to generalize, whereas curriculum-based strategies remained robust. Among these, HICO yielded the most consistent and well-structured parameter solutions. Moreover, models based on HICO provided the strongest and most reliable predictions of cognitive and psychopathology targets spanning multiple domains, including fluid reasoning ability (as measured using the Penn Matrix Test \cite{Moore2015}), behavioral problems related to internalization and externalization, as measured using the Achenbach Adult Self Report \cite{ACHENBACH20174}. These results indicate that incorporating domain-informed structure through evolutionary curricula fundamentally alters search dynamics, providing a principled mechanism for scaling high-dimensional brain models while yielding more stable, generalizable, and behaviorally informative solutions than flat optimization.

\section{Background}
\label{sec:background}

The following background situates the proposed HICO within prior work on whole-brain modeling and evolutionary search.

\subsection{Whole-Brain Biophysical Models}

A central goal of biophysical brain modeling is to explain how distributed functional dynamics emerge from the coupling of structural connectivity and local neural processes \cite{Deco2011,Breakspear2017}. Popular frameworks include neural mass, neural field, and Dynamic Mean Field (DMF) approaches. The resulting models can generate resting-state activity that reproduces key empirical signatures, including functional connectivity, dynamic functional connectivity, and metastability \cite{Deco2009,Cabral2014}.

A canonical study by \citet{Honey2009} showed that structural connectivity shapes—but does not fully determine—functional connectivity, motivating the use of computational models to explore this mapping. Subsequent work has characterized how factors such as excitation–inhibition balance, global coupling strength, conduction delays, and noise influence emergent functional patterns \cite{Deco2011,Breakspear2017}. The DMF model used in this paper is one such framework, balancing biological plausibility with computational tractability \cite{Deco2013}.

\subsection{Challenges in Parameter Optimization}

Increasing biological realism in large-scale brain models comes at the cost of sharply increased number of parameters. Strong nonlinear interactions between parameters give rise to complex, multimodal, and highly non-convex optimization landscapes. As a result, exhaustive search is infeasible, and gradient-based methods are often unsuitable due to non-differentiability, stochasticity, and numerical instability.

A central challenge is \emph{degeneracy}: many distinct parameter configurations may yield similar emergent dynamics \cite{Achard2006}. While such degeneracy reflects biological robustness, it complicates inference and increases the risk of overfitting idiosyncratic subject-level patterns. Steering optimization toward solutions that generalize across subjects and remain dynamically stable is therefore an important open research question.

\subsection{Cortical Hierarchy and Training Curricula}

For analysis, the brain is often parcellated into RSNs, i.e.\ functional subdivisions that are well-established in the literature \citep{Yeo2011}.  Recently, \citet{Margulies2016} identified a dominant gradient ranging from unimodal (sensorimotor) to transmodal (default mode) RSNs, thereby providing a macroscale order for cortical organization. 

We hypothesize that this organization of RSNs provide a natural basis for structured optimization.  Thus, suggesting  that parameters governing sensory-motor dynamics may be more foundational than those governing higher-order associative systems, or vice-versa. Motivated by this hierarchy, it is possible to construct phased and reverse-phased optimization curricula that introduce RSN-specific parameters in different temporal orders. This design makes it possible to test whether the order of parameter introduction acts as an inductive bias that shapes generalization and cross-subject stability.

Overall, HICO can be positioned as a domain-structured curriculum for EC that is distinct from (i) complexity-growth curricula (e.g., NEAT-style complexification), and (ii) permanent variable decomposition (e.g., cooperative coevolution). Its novelty is to use a biologically grounded hierarchy to define a \emph{phase ordering} over pre-existing parameter blocks in a whole-brain dynamical model, in order to generalize across subjects better and predict behavior more accurately, rather than simply fit to the data.

\subsection{Curricula and Staged Search in Evolutionary Computation}

The use of \emph{curricula} in EC often refers to deliberately shaping the search process over time so that later stages become feasible only after earlier structure has been established. One prominent line is \emph{incremental evolution} or \emph{complexification}, where problem difficulty or model complexity is increased gradually, e.g. by adding degrees of freedom, structure, or task demands \cite{Stanley2002NEAT}. The idea is to  help evolution avoid brittle local optima early on and discover scalable representations. Related ideas appear in staged or progressive training regimes in evolutionary robotics and neuroevolution, where simpler components are learned before integrating higher-level behaviors, effectively regularizing the search trajectory through time~\cite{Gomez1997}.

HICO differs from such approaches in an important way: rather than increasing task difficulty or expanding representation size, it imposes a \emph{domain-derived ordering} over parameter blocks that already exist in the model. In whole-brain DMF models, parameters are naturally grouped by RSNs and linked to known cortical gradients; HICO leverages this structure by prioritizing higher-level organizational degrees of freedom before specializing RSN-specific parameters. This makes HICO closer to \emph{structured staged optimization} than to classic ``easy-to-hard'' curricula.

\subsection{Decomposition, Cooperative Coevolution, and Multiobjective Optimization}
Another major family of EC metaheuristics addresses high dimensional search by decomposing variables into subcomponents and optimizing them with limited coordination, most notably \emph{cooperative coevolution} \cite{Potter2000CCEA}. Cooperative coevolution and its many variants explicitly model the fact that different subsets of parameters may interact at different strengths, and they often rely on collaboration schemes to assemble partial solutions into a full candidate. Modern surveys emphasize decomposition design and variable-interaction structure learning as central to performance in high-dimensional domains \cite{Ma2019CCEA}.

HICO is conceptually aligned with this literature in that it treats the parameter vector as \emph{structured}, but it differs operationally: rather than permanently decomposing the problem into concurrently evolving subpopulations, HICO introduces a \emph{temporal decomposition} (a curriculum) that controls when each RSN-specific block is allowed to vary. The curriculum preserves a single coherent model throughout optimization while still exploiting modular structure to reduce destructive interference between strongly coupled parameter groups.

One way to obtain solutions that not only are a good fit with data but also generalize and predict performance would be to use measurements of those qualities as secondary objectives. However, those objectives may not be known ahead of time. The model may be used later to diagnose disorders, predict measures of cognitive abilities, or propensity for psychopathology. The goal of HICO is to guide evolution towards useful models regardless of what other metrics may later be of interest. The rationale for not adopting a multiobjective formulation in this setting is discussed in Appendix~\ref{app:clarifications}, where the distinction between model fitting and downstream evaluation is clarified.


\subsection{Evolutionary Algorithms in Neuroscience}

Evolutionary algorithms (EAs) maintain a population of candidate solutions that evolve under selection, mutation, and recombination \cite{Holland1992}. They are especially effective for high-dimensional, non-differentiable problems with complex fitness landscapes. Early applications demonstrated their value in fitting single-neuron conductance parameters \cite{Vanier1999,Keren2005}. Subsequently they were used to fit detailed Purkinje cell models, revealing high-dimensional manifolds of viable solutions \citep{Achard2006}. Further, using the multi-objective evolutionary approach, it was possible to match multiple electrophysiological features simultaneously \citep{Druckmann2007}. At larger scales, population-based evolutionary optimization has been used to tune RSNs and whole-brain models \cite{vanGeit2016,Maile2019}.

Among the many variants of EAs, covariance-based strategies such as CMA-ES have proven highly effective in moderate dimensional settings \cite{Hansen2016}. However, they are likely to be less effective in the high-dimensional spaces such as those resulting from neuroscience applications. For this reason, this paper relies on the traditional genetic algorithms: they are scalable even with future extensions to large-scale models. Further, they are flexible, and can thus readily accommodate structured interventions such as the curriculum-based parameter scheduling. In any case, the experiments in this paper do not rely on any optimization method in particular; the goal is to evaluate the effect of curriculum as an inductive bias in general.

\section{Method}
\label{sec:method}

This section describes the data, modeling framework, optimization objectives, and evolutionary strategies used to compare homogeneous and heterogeneous whole-brain DMF models under different optimization curricula.

\begin{figure}
    \centering
    \includegraphics[width=0.80\linewidth]{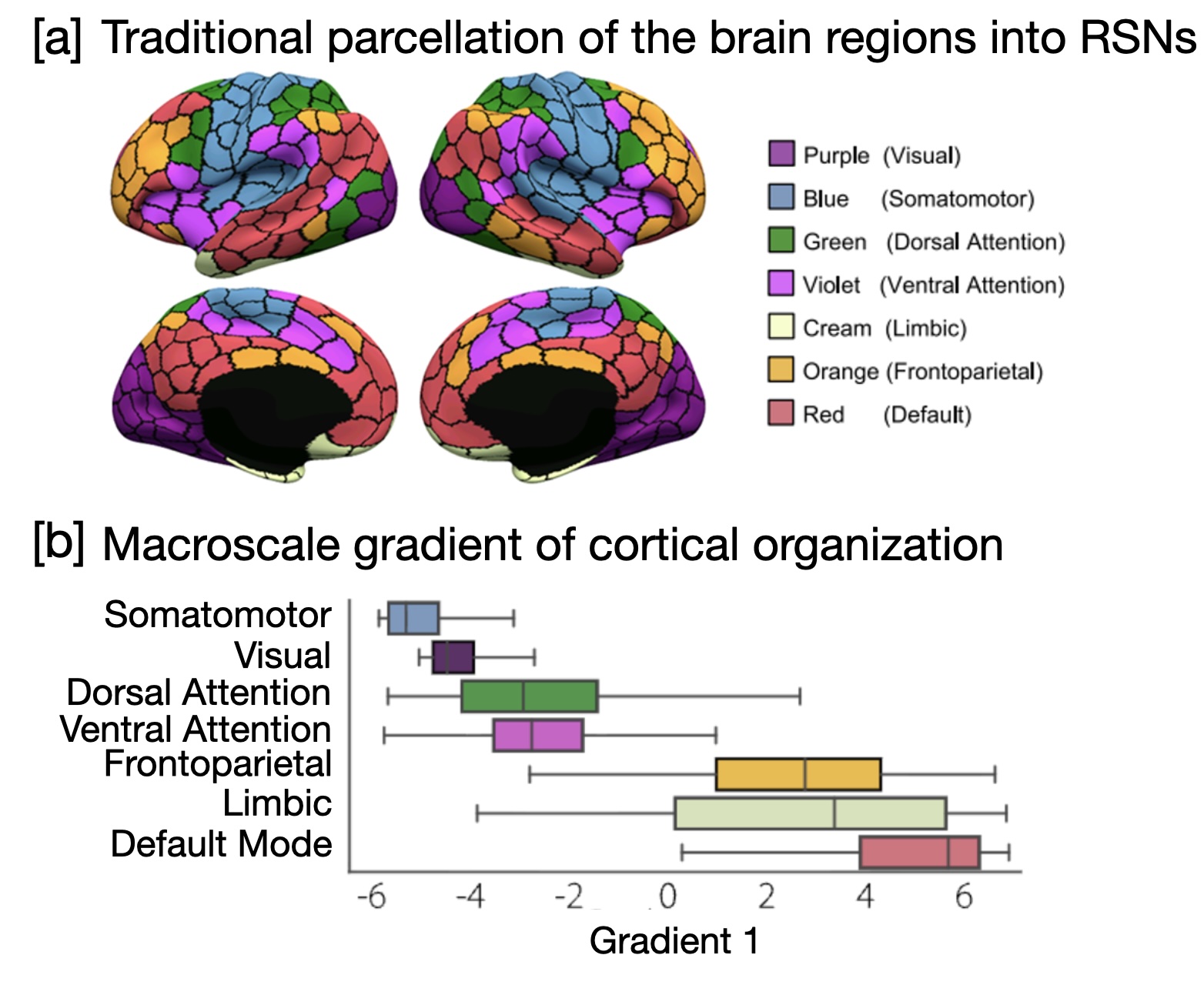}
    \vspace*{-2ex}
    \caption{\footnotesize 
    Large-scale cortical organization used to define RSN-specific parameter blocks.
    \textbf{(a)} Surface renderings of a 400-region cortical parcellation mapped to the seven canonical RSNs of \citet{Yeo2011}, shown for lateral and medial views of both hemispheres.
    Colors denote Visual, Somatomotor, Dorsal Attention, Ventral Attention, Limbic, Frontoparietal Control, and Default Mode RSNs.
    \textbf{(b)} Distribution of parcels along the principal macroscale gradient of cortical organization (Gradient~1), ordered from unimodal sensory–motor systems to transmodal association cortex.
    This gradient defines the hierarchical ordering used to construct curriculum phases in the proposed optimization framework.}
    \Description{}
    \label{fig:parcellation}
\end{figure}

\subsection{Data and Parcellation}

Resting-state functional Magnetic Resonance Imaging (rs-fMRI) and diffusion-weighted imaging (DWI) data were obtained from 100 unrelated subjects from the Human Connectome Project (HCP). The cerebral cortex was parcellated into 400 regions using the \citet{Schaefer2018} atlas, with parcels assigned to seven canonical RSNs according to the \citet{Yeo2011} parcellation (Fig.~\ref{fig:parcellation}). Subject-specific structural connectivity (SC) matrices were derived from tractography of the DWI data, and empirical functional connectivity (FC) matrices were computed as Pearson correlations of regional Blood-Oxygen-Level–Dependent (BOLD) rs-fMRI time series.

\subsection{Dynamic Mean Field Model}

Each cortical region was modeled as a coupled excitatory--inhibitory neural mass using the Dynamic Mean Field (DMF) framework \cite{Deco2011,Deco2009,Breakspear2017}. 
The DMF model provides a mesoscale approximation to large populations of spiking neurons; it was shown to reproduce key properties of resting-state dynamics when driven by subject-specific structural connectivity \cite{Deco2009,Murray2014}.

Let $S_i(t)$ denote the excitatory synaptic gating variable for region $i$. 
The excitatory population dynamics are governed by
\begin{equation}
\tau_E \frac{dS_i}{dt}
= -S_i + (1 - S_i)\,\gamma\,H\!\left(I_i(t)\right),
\end{equation}
where $\tau_E$ is the excitatory time constant, $\gamma$ controls synaptic gain, and $H(\cdot)$ is a sigmoidal firing-rate function. 
Long-range coupling between regions is defined as
\begin{equation}
I_i(t) = I_b + w_E S_i(t) - w_I S^I_i(t)
+ G \sum_{j} \mathrm{SC}_{ij} S_j(t - d_{ij}),
\end{equation}
where $I_b$ is a background current, $w_E$ and $w_I$ denote local excitatory and inhibitory coupling strengths, $S^I_i(t)$ is the inhibitory gating variable, $G$ is the global coupling parameter, $\mathrm{SC}_{ij}$ is the structural connectivity weight from region $j$ to $i$, and $d_{ij}$ represents conduction delays. 
Canonical DMF constants were adopted from prior work \cite{Deco2009}, except for the parameters optimized in this study.
See Appendix~\ref{app:dmf_parameters} for the full specification of DMF variables and constants.
The homogeneous model comprised 20 shared parameters applied uniformly across all regions. 
The heterogeneous formulation comprised 140 parameters, corresponding to seven RSNs with independent copies of the same 20 parameters. 
See Appendix~\ref{app:rsn_params_table} for the full parameter specification.
In both cases, the DMF equations define a stochastic dynamical system that generates region-wise neural activity time series whose statistical structure depends on the parameter vector $\Theta$.

\subsection{Simulation and Objective Function}

For a candidate parameter vector $\Theta$, the DMF system defined above was simulated for five minutes of biological time using an Euler--Maruyama integration scheme~\cite{Higham2001} with a step size of 0.1\,ms. 
The resulting excitatory synaptic activity $S_i(t)$ was converted into BOLD signals using the Balloon--Windkessel hemodynamic model \cite{Friston2003}. 
Functional connectivity matrices $\widehat{\mathrm{FC}}(\Theta)$ were then computed from the simulated BOLD time series by taking pairwise Pearson correlations between all cortical regions, in direct analogy to the empirical FC derived from resting-state fMRI.
See Appendix~\ref{app:pipeline} for implementation details.
The DMF equations thus induced, via simulation and hemodynamic transformation, a mapping
\[
\Theta \;\longmapsto\; \widehat{\mathrm{FC}}(\Theta),
\]
from model parameters to predicted functional connectivity. 
Parameter fitting was formulated as an inverse problem: finding $\Theta$ such that the FC generated by the DMF dynamics matches the empirical FC as closely as possible.

The optimization objective minimized the discrepancy between empirical and simulated FC:
\begin{equation}
\label{eq:dmf_objective}
\mathcal{L}(\Theta)
= 1 - \mathrm{corr}\!\left(
\mathrm{vec}(\mathrm{FC}),\,
\mathrm{vec}\!\left(\widehat{\mathrm{FC}}(\Theta)\right)
\right),
\end{equation}
where $\mathrm{vec}(\cdot)$ extracts the upper triangular elements of the FC matrices. 
Equation~\eqref{eq:dmf_objective} directly links the DMF dynamics to the optimization criterion: the DMF equations generate neural activity and BOLD time series, these time series define $\widehat{\mathrm{FC}}(\Theta)$, and the loss function measures how well the resulting large-scale correlation structure matches empirical resting-state connectivity.
See Appendix~\ref{app:fitness} for details of the fitness computation.

This FC-based objective follows standard practice in whole-brain modeling, where functional connectivity serves as a compact summary of long-range statistical dependencies induced by the underlying neural dynamics~\cite{Deco2009,Murray2014,Maile2019}. 
Minimizing $\mathcal{L}(\Theta)$ therefore corresponds to calibrating the biophysical parameters of the DMF model so that its emergent RSN-level behavior reproduces the observed functional organization of the human brain.

\subsection{Optimization Curricula}
\label{sec:optimization_curricula}

This study compares multiple evolutionary optimization curricula that differ
only in the temporal ordering by which parameter subsets are optimized during
search.

\paragraph{Homogeneous Baseline}
A single set of 20 parameters was used for all RSNs. Thus, this baseline assumes that biophysical dynamics are homogeneous throughout the brain.

\paragraph{Heterogeneous Optimization}
All 140 RSN-specific parameters (20 parameters for each of the seven RSNs) were
optimized simultaneously from the outset, without any curriculum structure.
This approach maximizes search dimensionality at all times and serves as a
flat baseline for assessing the impact of curriculum structure.

\paragraph{Hierarchy-Informed Curriculum Optimization (HICO)}
Guided by the principal cortical gradient described by Margulies et al.\
\cite{Margulies2016}, parameters were optimized sequentially according to the
following phases:

\begin{itemize}
    \item \textbf{Phase I — Global Phase.}
    This phase establishes the homogeneous baseline as the starting point. Further optimizaiton is thus restricted to a low-dimensional, dynamically coherent subspace, which reduces the unstable fixed points and improves Lyapunov stability~\cite{Deco2009}.
    \item \textbf{Phase II — Transmodal core (Default Mode and Limbic RSNs).}
    \item \textbf{Phase III — Frontoparietal control RSN.}
    \item \textbf{Phase IV — Attention RSNs (DorsAttn and SalVentAttn).}
    \item \textbf{Phase V — Visual RSN (VIS).}
    \item \textbf{Phase VI — Somatomotor RSN (SomMot).}
\end{itemize}

Within each phase, only the parameters associated with the active hierarchical
segment were allowed to vary, while parameters optimized in earlier phases were
held fixed. The total generation budget was divided evenly across phases, such
that all curricula were matched for overall computational cost.

\paragraph{Reverse-Phased Curriculum}
The reverse-phased curriculum retained an identical Phase~I (global parameters)
but executed the remaining five hierarchy-specific phases in reverse order,
beginning with the transmodal core and progressing toward unimodal sensory
systems. Thus, this curriculum tested whether the direction of hierarchical parameter
release influences optimization outcomes.

\paragraph{Shuffled Curriculum}
In the shuffled curriculum, Phase~I (global parameters) was again identical to
HICO, while the remaining five hierarchy-specific phases were randomly permuted
for each subject. This control preserved the phased structure of optimization
while removing any alignment with the cortical hierarchy.

\subsection{Evolutionary Optimization Method}
Model parameters were optimized using an elitist genetic algorithm (GA). Each candidate solution is a fixed-length parameter vector:
$d=20$ for the homogeneous model and $d=140$ for heterogeneous models.

To ensure comparable search resolution across parameters with different physical units, each parameter was represented in a normalized coordinate in $[0,1]$ with 1000-step precision and mapped via an affine transform to its literature-supported valid range, as established in prior DMF modeling studies \cite{WongWang2006,Deco2009,Breakspear2017}. This representation yields an effective discrete search space of $10^{60}$ possible configurations for the homogeneous model and $10^{420}$ for the heterogeneous formulation. This scale makes the evolutionary approach a good choice for optimizing these models.

Although Eq.~\eqref{eq:dmf_objective} is written as a loss function, evolutionary optimization operated by \emph{maximizing} the corresponding fitness, defined as the Pearson correlation between empirical and simulated functional connectivity matrices (i.e., $ 1 - \mathcal{L}(\Theta)$). Throughout this work, the terms \emph{fitness} and \emph{score} refer to this correlation value, with higher scores indicating closer agreement between simulated and empirical FC.

Across all optimization curricula, the GA configuration was held constant as
\begin{itemize}
    \item \textbf{Population size:} $N=100$ individuals per generation.
    \item \textbf{Elitism:} the top $E=20$ individuals (highest fitness) were carried
          forward unchanged to the next generation.
    \item \textbf{Parent selection:} tournament selection with tournament size $k=3$.
    \item \textbf{Variation operators:} uniform crossover and per-gene mutation.
    \item \textbf{Mutation rate:} $p_{\mathrm{mut}}=0.1$ (probability of mutating each gene).
    \item \textbf{Generation budget:} a fixed total number of generations $G_{\mathrm{total}}=120$
          was used for all methods to ensure compute-matched comparisons.
\end{itemize}

For curriculum-based strategies, \(G_{\mathrm{total}}\) was divided evenly across six hierarchy-defined phases (as was described in Section~\ref{sec:optimization_curricula}).
Within each phase, only the parameter subset corresponding to the active RSN was allowed to vary, while parameters optimized in prior phases were held fixed.
This schedule yielded an equal overall computational budget while controlling the effective search dimensionality over time.

All experiments were parallelized using a hub-and-spoke distributed evolutionary architecture with four concurrent evolutionary engines \cite{Shahrzad2023}.
In this setup, multiple worker spokes evaluate candidate solutions in parallel and synchronize with a central hub that maintains population state and applies selection and variation. This design increases throughput without changing the underlying optimization dynamics, and ensures identical GA settings across all methods. It thus enables a controlled comparison that isolates the impact of curriculum on the optimization process.

\section{Results}
\label{sec:results}

This section evaluates how different evolutionary optimization strategies influence model fit to individual subjects, cross-subject generalization, parameter geometry, and behavioral relevance in whole-brain DMF models. First, the approaches are evaluated in fitting to individual subjects in comparison against a homogeneous baseline. Second, generalization to new subjects is tested in leave-one-out (LOO) experiments. Third, to elucidate the mechanisms underlying observed failures and successes, the geometry of learned parameter distributions is examined using low-dimensional embeddings.
Fourth, behavioral relevance of optimized parameters is assessed through RSN-level prediction analysis, and fifth, through permutation-calibrated statistical testing. Together, the results demonstrate that while increased model dimensionality improves within-subject fit, the \emph{temporal structure} imposed by HICO is critical for achieving stable, generalizable, and behaviorally informative solutions.

\begin{figure}[t]
  \centering
  \includegraphics[width=0.8\linewidth]{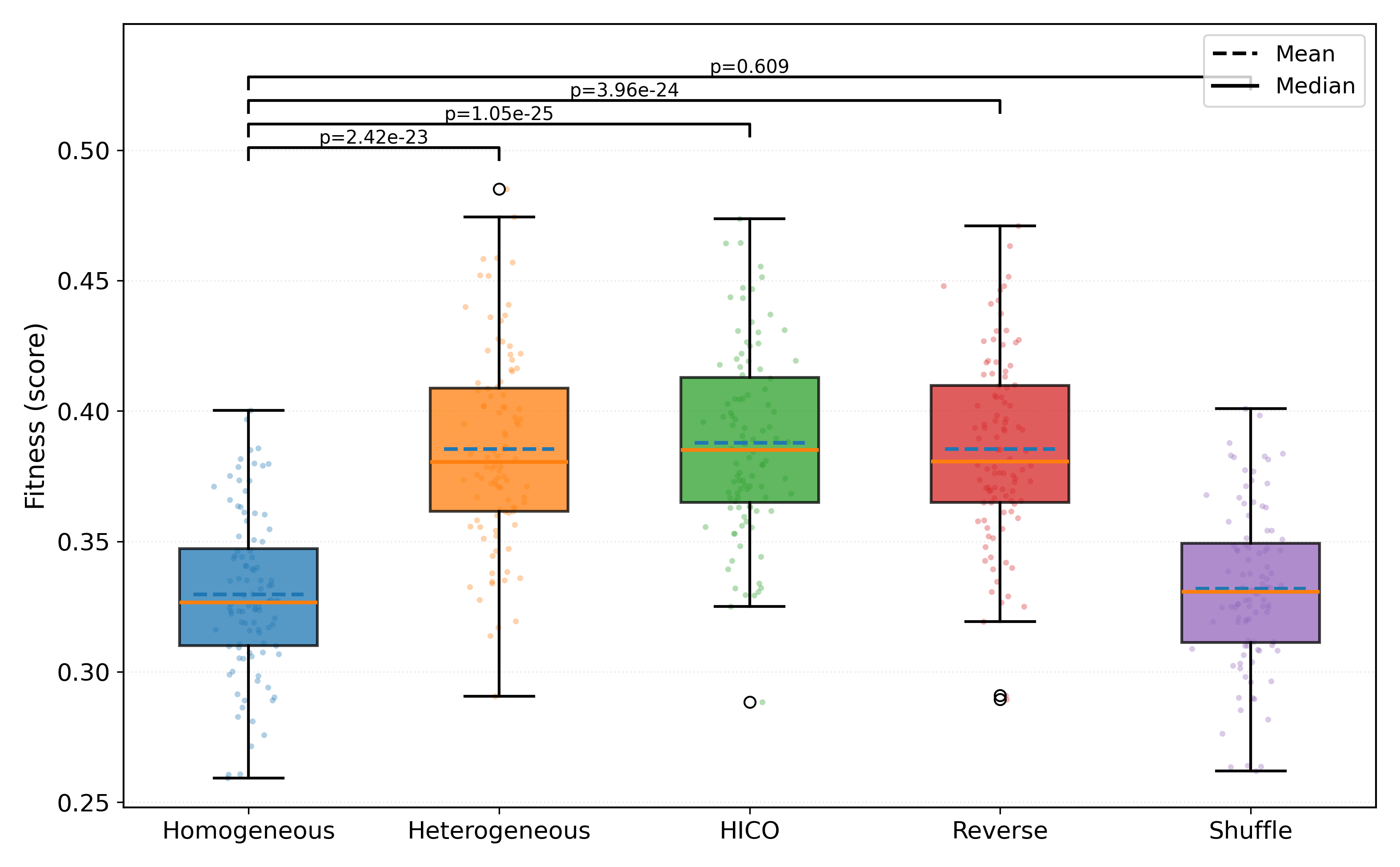}
  \vspace*{-2ex}
  \caption{\footnotesize 
  Distribution of fitness scores across optimization strategies when models were optimized separately for each individual subject. The points represent subjects and the boxes show the interquartile range (IQR) with median (solid line) and mean (dashed line).
  Unpaired statistical tests against the homogeneous baseline indicate that the heterogeneous, HICO, and reverse curricula achieve significantly higher fitness, demonstrating the advantage of increased model dimensionality. In contrast, the shuffled curriculum fails to realize this benefit, suggesting that the curriculum needs to be systematic to take advantage of such dimensionality.}
  \Description{}
  \label{fig:fitness_distribution_baseline}
\end{figure}

\subsection{Fitting to Individual Subjects}

Figure~\ref{fig:fitness_distribution_baseline} summarizes the distribution of fitness scores obtained under each optimization approach when parameters were optimized separately for each individual subject.

Relative to the homogeneous baseline, flat heterogeneous optimization,  HICO, and the reverse-phased curriculum all exhibit upward shifts in both median and mean fitness. Paired t-tests against the homogeneous baseline indicate that these improvements are highly significant (Heterogeneous: \(p=2.42\times10^{-23}\); Phased: \(p=1.05\times10^{-25}\); Reverse: \(p=3.96\times10^{-24}\)), confirming that introducing RSN-specific parameters substantially improves fit to individual subjects. In contrast, the shuffled curriculum shows substantial overlap with the homogeneous distribution and does not achieve a statistically significant improvement over baseline (\(p=0.609\)).

These results establish two key points. First, increasing model complexity by moving from homogeneous to heterogeneous parameterizations yields a robust and statistically significant gain in fitting to individual subjects. Second, imposing a biologically informed curriculum does not compromise this fit: both HICO and reverse-phased optimization achieve fitness levels comparable to or exceeding those of flat heterogeneous optimization. This finding provides an essential baseline for subsequent analyses, demonstrating that curriculum-based strategies retain strong within-subject performance while enabling further investigation of
their effects on generalization, stability, and behavioral relevance.

\begin{figure}[t]
  \centering
  \includegraphics[width=0.8\linewidth]{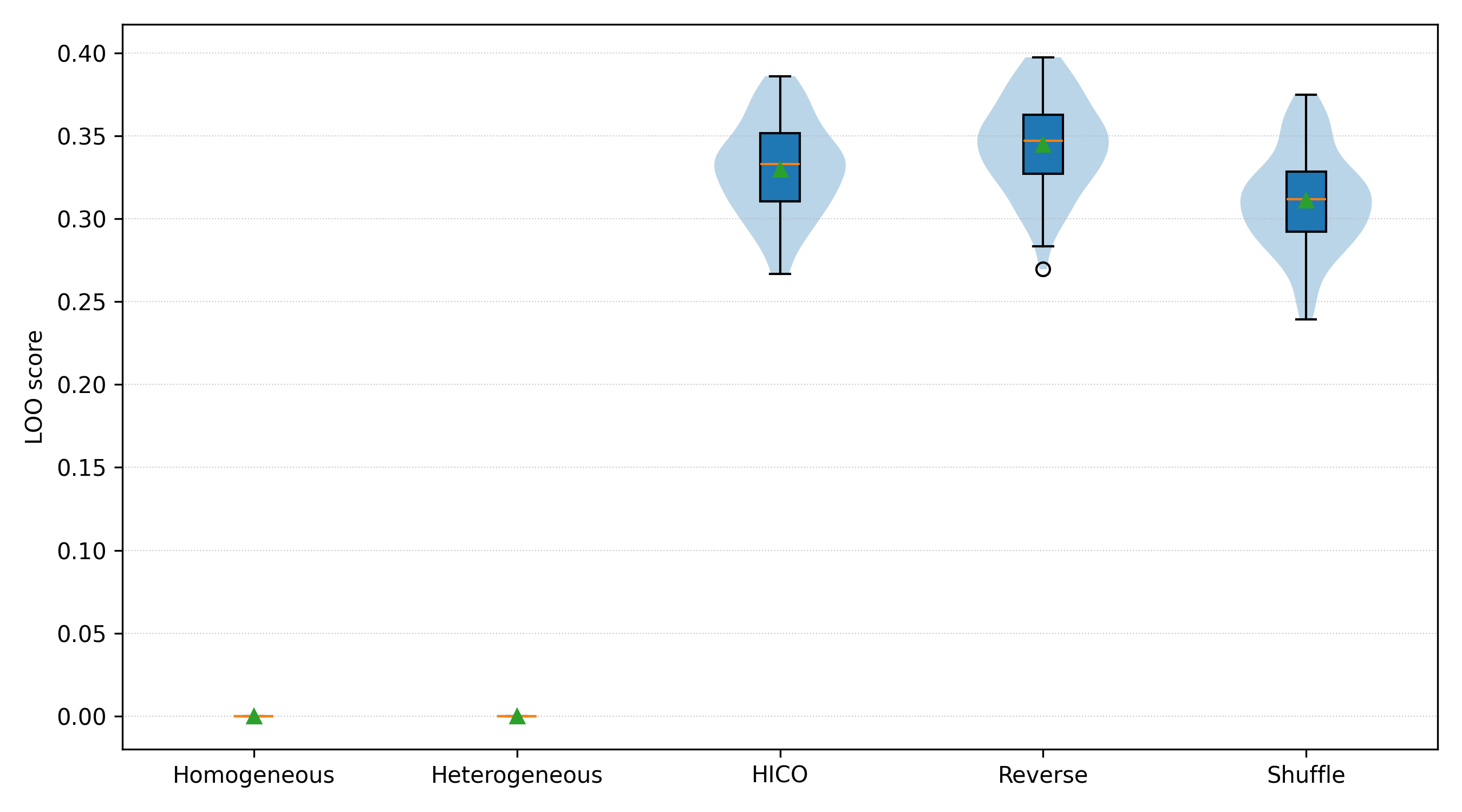}
  \vspace*{-2ex}
  \caption{\footnotesize Leave-one-out (LOO) fitness distributions across optimization strategies. Violin plots show the distribution of LOO fitness scores across subjects, with embedded boxplots indicating median and interquartile range; triangles denote means. Homogeneous and flat heterogeneous strategies collapse to zero LOO fitness across the board, reflecting dynamical instability LOO fitness calculation. In contrast, curriculum-based strategies—particularly HICO and reverse-phased curricula—maintain robust, non-degenerate LOO performance, indicating improved cross-subject generalization.}
  \Description{}
  \label{fig:loo_boxplot}
\end{figure}

\subsection{Generalization across Subjects}
\label{sc:generalization}

Cross-subject generalization was evaluated using a leave-one-out (LOO) parameter averaging based on a two-sided trimmed mean estimator ($p=0.1$). The trimmed mean provides a robust estimate of central tendency by discarding extreme values from both tails of the parameter distribution, reducing sensitivity to outlier solutions while preserving the dominant structure shared across subjects \cite{Huber2009}. For each left-out subject, parameters fitted on the remaining subjects were averaged and evaluated against the empirical data of the left-out subject.

Figure~\ref{fig:loo_boxplot} reveals a divergence between flat and curriculum-based strategies under this evaluation. Despite achieving competitive fitness in fitting to individual subjects, both the homogeneous baseline and flat heterogeneous optimization frequently produced averaged parameter vectors with LOO fitness scores at zero. This collapse indicates that the averaged parameters lie in dynamically unstable regimes of the DMF model, consistent with failure to solve the associated Lyapunov equations and resulting in degenerate dynamics. In contrast, all curriculum-based strategies consistently avoided such instability, yielding no averaged parameter vectors with degenerate or unstable dynamics.

These findings identify an interesting generalization failure mode for unconstrained high-dimensional optimization: parameter configurations that fit individual subjects well may not survive averaging across subjects. Curriculum-guided optimization meets this challenge by constraining search trajectories toward dynamically coherent regions of parameter space that remain stable when averaged across subjects.

\begin{figure}[t]
  \centering
  \includegraphics[width=0.8\linewidth]{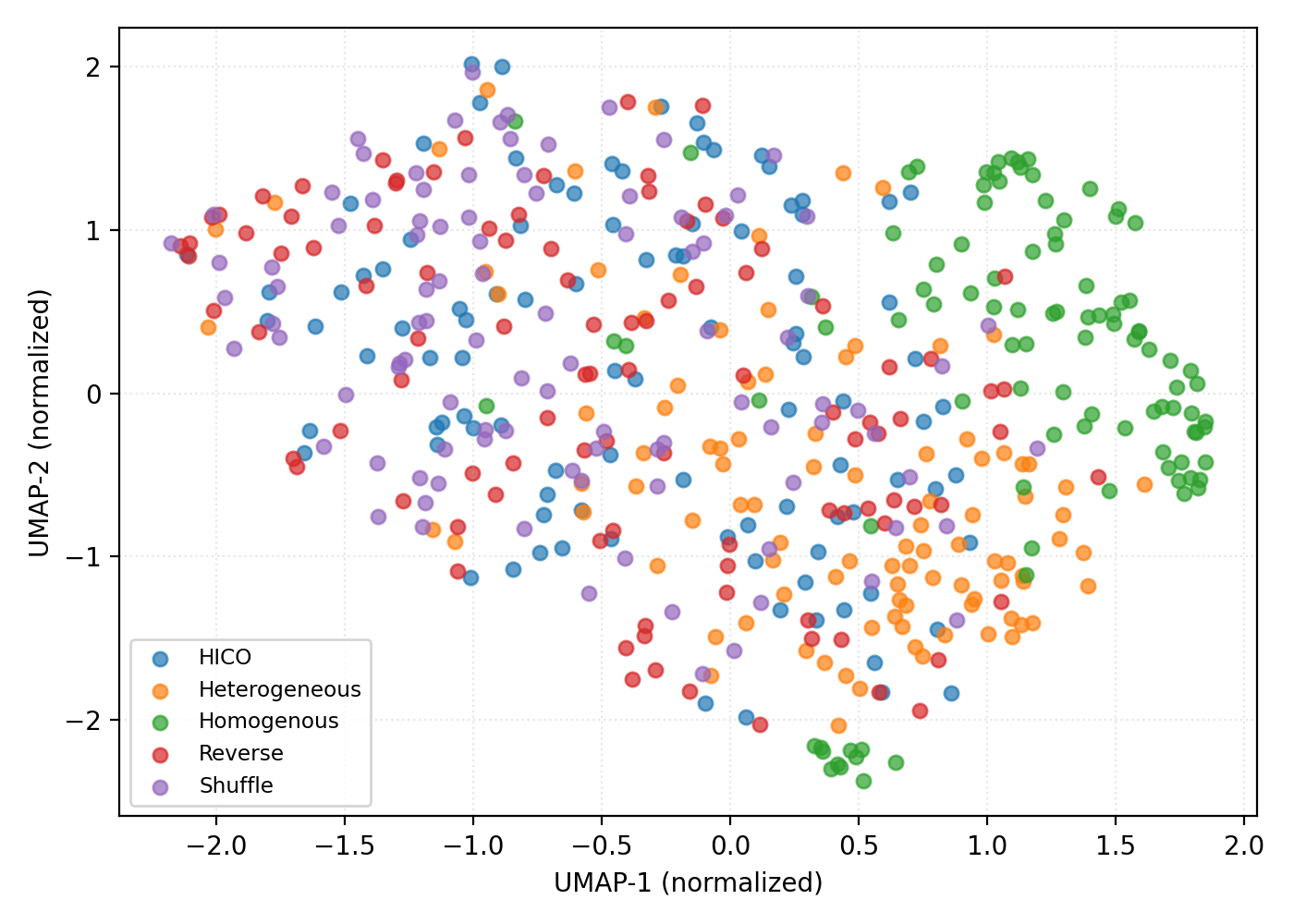}
  \vspace*{-2ex}
  \caption{\footnotesize UMAP embedding of subject-level parameter vectors (averaged by parameter type). Each point is a subject, colored according to the approach used. Curriculum-based methods (HICO, Reverse, Shuffled) occupy a large overlapping region of parameter space, whereas homogeneous and heterogeneous approaches form distinct, focused clusters shifted away from this region. As seen in Fig.~\ref{fig:loo_boxplot}, the generalization properties are very different in these two areas.}
  \Description{}
  \label{fig:param-umap}
\end{figure}

\subsection{Geometry of Solutions}

To probe how different optimization strategies shape the geometry of the learned parameter space, subject-level parameter vectors were embedded into two dimensions using Uniform Manifold Approximation and Projection (UMAP)~\cite{McInnes2018UMAP}. UMAP preserves local neighborhood structure while providing a qualitative view of large-scale organization, making it possible to see how the different methods populate parameter space.

Figure~\ref{fig:param-umap} reveals marked differences between flat and curriculum-based strategies. Homogeneous and flat heterogeneous optimizations form relatively compact and well-separated clusters, indicating that these methods converged toward narrow regions of parameter space across subjects. Such concentration suggests that the parameter manifolds are fragile or degenerate: small perturbations may push the system outside dynamically stable regimes.

In contrast, HICO and other phased curricula (Reverse-Phased and Shuffled) produce embeddings that are more broadly distributed and substantially overlapping. Rather than collapsing onto a single narrow cluster, these methods populate an extended region of parameter space, suggesting that curriculum-based optimization facilitates access to a wider set of dynamically viable solutions. This organization is robust under cross-subject parameter averaging, as illustrated in Section~\ref{sc:generalization}.

\subsection{Predicting Behavior}
\label{sc:behavior}

Behavioral relevance of optimized DMF parameters was assessed by predicting three behavioral targets from RSN-level parameter summaries: fluid reasoning ability (PMAT24\_A\_CR) and behavioral problems related to either internalization (inwardly directed affective and somatic symptoms, ASR\_Intn\_Raw) or externalization (outwardly directed behavioral tendencies, ASR\_Extn\_Raw). For each approach, parameters for the seven RSNs were averaged into a single 20 dimensional vector and ridge regression models were fit to each behavioral target separately. Predictive performance was then quantified using the coefficient of determination ($R^{2}$), which measures the proportion of behavioral variance explained by the model, relative to a baseline that predicts the target's mean across subjects.

\begin{figure*}[t]
  \centering
  \includegraphics[width=0.8\textwidth]{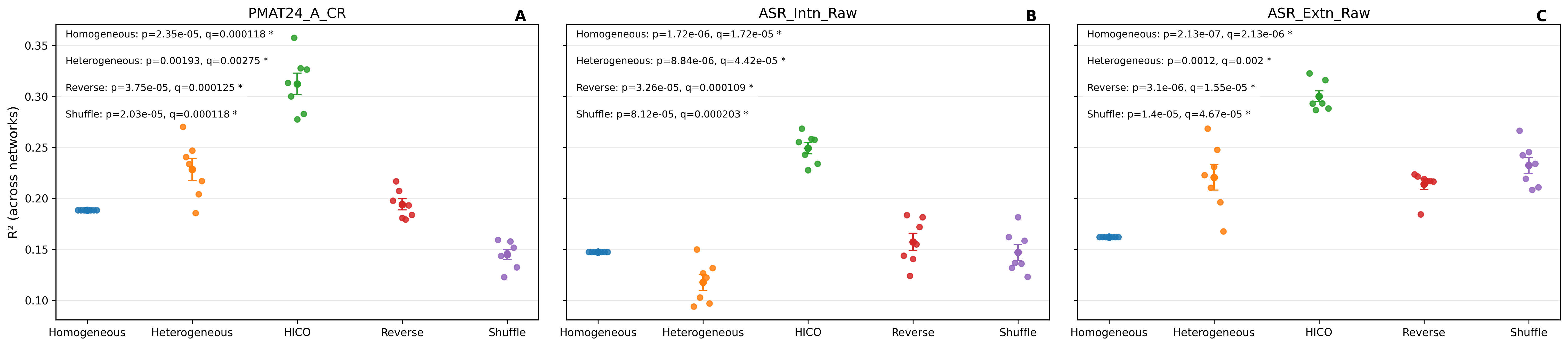}
  \vspace*{-2ex}
  \caption{\footnotesize Predicting behavior based on solutions obtained by the different optimization strategies. Each subplot reports ridge regression $R^{2}$ values (across RSNs) for a different behavioral target: (A) fluid reasoning ability (PMAT24\_A\_CR), (B) inwardly directed affective and somatic symptoms (ASR\_Intn\_Raw), and (C) outwardly directed behavioral tendencies (ASR\_Extn\_Raw). For each optimization strategy, colored points show the $R^{2}$ values obtained for the seven individual RSNs, and error bars indicate the mean $\pm$ standard deviation across RSNs. Reported $p$-values and $q$-values correspond to permutation-based significance testing with False Detection Rate (FDR) correction. Across all targets, only HICO produced solutions that encode behaviorally relevant information.}
  \Description{}
  \label{fig:fig_methods_by_target}
\end{figure*}

Figure~\ref{fig:fig_methods_by_target} shows a consistent advantage for HICO across all behavioral domains. For fluid reasoning (Fig.~\ref{fig:fig_methods_by_target}A), HICO achieved the highest $R^{2}$ values across all the strategies, substantially exceeding both the homogeneous baseline and the flat heterogeneous optimization, with statistically significant improvements after multiple-comparison correction. A similar pattern was observed for behavioral problem scales, internalization scale - ASR\_Intn\_Raw (Fig.~\ref{fig:fig_methods_by_target}B) and externalization scale - ASR\_Extn\_Raw (Fig.~\ref{fig:fig_methods_by_target}C).

Across all three behavioral targets, reverse-phased curricula achieved intermediate performance, whereas shuffled curricula had lower and more variable $R^{2}$ values. These results indicate that merely increasing parameter dimensionality is insufficient to recover behaviorally meaningful representations. Instead, the ordering imposed by hierarchy-informed curricula plays a critical role in shaping parameter configurations that capture individual differences relevant to cognition and behavior. Together, these findings demonstrate that HICO not only stabilizes cross-subject generalization but also results in solutions that can be interpreted behaviorally.

\subsection{Reliability of Predictions}
\label{sc:permutation}

To assess whether the observed associations between optimized DMF parameters and behavioral measures exceeded chance levels, prediction performance was evaluated using a permutation-calibrated significance framework. More specifically, the $R^2$ values obtained in Section~\ref{sc:behavior} were evaluated against an empirical null distribution generated by repeatedly permuting the behavioral labels across subjects.

\begin{figure*}[t]
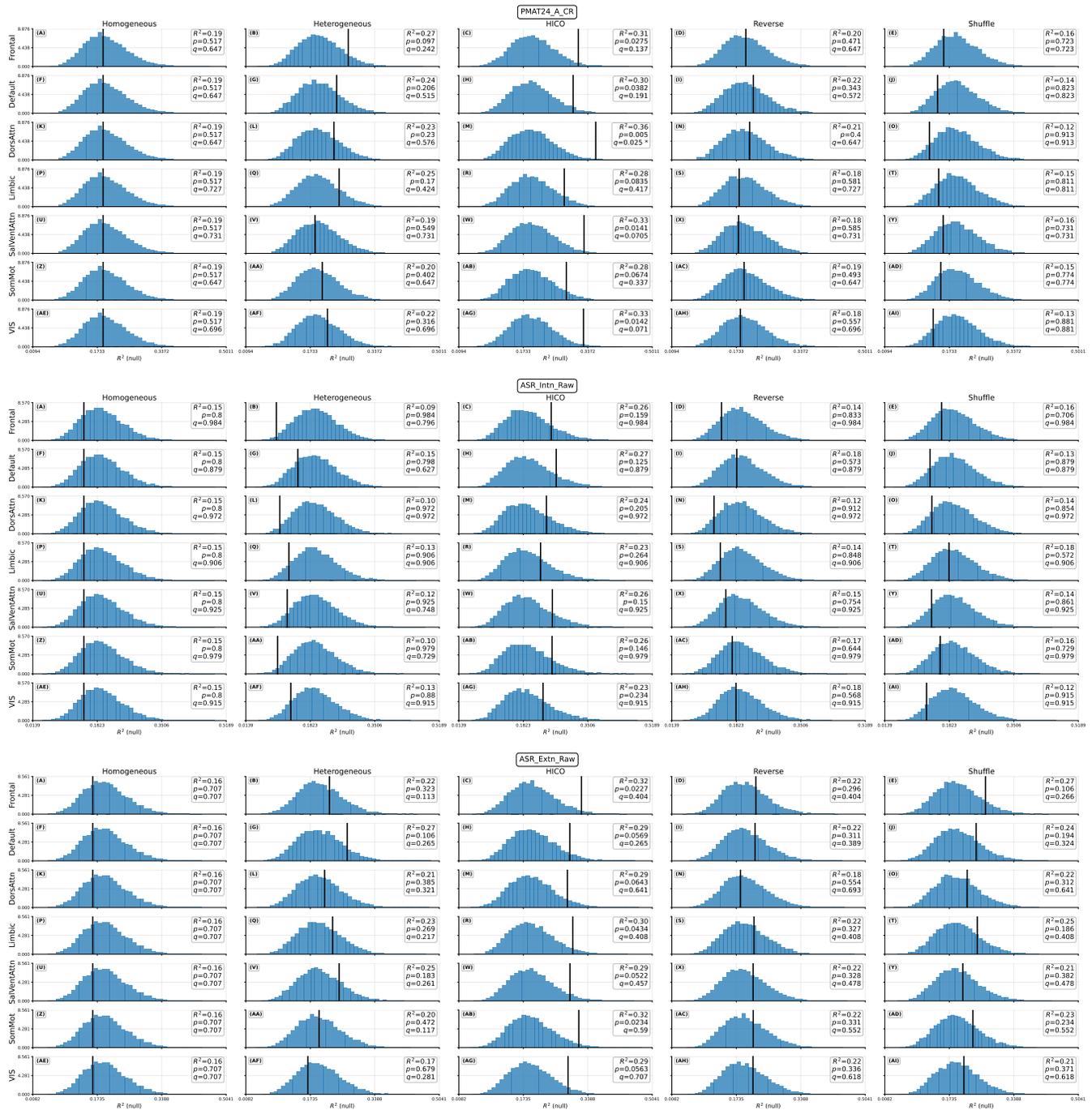

  \centering
  \includegraphics[height=0.25\textheight]{figures/localization_rowwise_PMAT24_A_CR_multi_panel.png}

  \vspace{0.5em}
  \includegraphics[height=0.25\textheight]{figures/localization_rowwise_ASR_Intn_Raw_multi_panel.png}

  \vspace{0.5em}
  \includegraphics[height=0.25\textheight]{figures/localization_rowwise_ASR_Extn_Raw_multi_panel.png}
  
  \vspace*{-2ex}
  \caption{\footnotesize 
  Localization of permutation-calibrated parameter--behavior associations across the different RSNs.
  Top: fluid reasoning ability (PMAT24\_A\_CR).
  Middle: inwardly directed behavioral problems (ASR\_Intn\_Raw).
  Bottom: outwardly directed behavioral problems (ASR\_Extn\_Raw).
  For each RSN and optimization approach, histograms show the null distribution of ridge regression $R^2$ obtained from 10{,}000 permutations of behavioral labels, while vertical lines indicate the observed $R^2$ from the true data.
  Reported $p$- and $q$-values quantify statistical significance relative to the permutation null. As indicated by the vertical lines, biology-informed optimization results in parameter solutions that consistently exceed chance levels and provide a strong foundation for predicting behavioral.
  }
  \Description{Permutation-calibrated RSN-wise behavioral prediction results across optimization strategies.}
  \label{fig:behavior_localization_all}
\end{figure*}

Figure~\ref{fig:behavior_localization_all} summarizes the resulting RSN-wise null distributions for the three behavioral targets. Within each panel, histograms represent the distribution of $R^2_{\mathrm{null}}$ values obtained from 10{,}000 permutations, while the vertical line indicates the observed $R^2_{\mathrm{true}}$ for the corresponding RSN and optimization approach. The accompanying annotations report the observed $R^2$ as well as permutation-derived $p$-values and false discovery rate–corrected $q$-values.

Across all three behavioral targets, HICO produced the most consistent and pronounced deviations from the null distribution. Multiple RSNs under HICO had $R^2$ values well in the upper tail of the permutation distribution. These effects are most robust for fluid reasoning ability. In contrast, other optimization strategies yielded $R^2$ values that fell near the center of the null distribution, indicating weak or unreliable parameter–behavior relationships.

Importantly, this permutation-based analysis complements the behavioral prediction results by localizing where among the RSNs a behaviorally meaningful signal emerges. The findings demonstrate that the superior predictive performance of HICO is not driven by diffuse, nonspecific effects, but rather by structured and statistically considerable associations within specific cortical systems. Together, these results indicate that solutions found by the biologically motivated curriculum approach predict behavior in a statistically reliable manner.

\section{Discussion and Future Work}
\label{sec:Discussion}

The experimental results demonstrate that the curricular structure of parameter optimization is crucial in constructing whole-brain biophysical models that generalize across subjects and can be used to predict behavior. Although evolutionary algorithms are designed to explore high-dimensional search spaces without explicit inductive biases, the order in which parameter subspaces are explored can strongly shape the resulting search trajectory. In Dynamic Mean Field (DMF) models, where nonlinear interactions give rise to emergent large-scale dynamics, curriculum structure acts as an effective constraint that promotes stability, cross-subject generalization, and behavioral relevance.

A natural next step is to scale this framework to larger and more diverse datasets. In particular, extending the analysis to the full Human Connectome Project (with $n \approx 1200$ subjects) will allow characterizing inter-individual variability more precisely, thus making behavior predictions more accurate. In parallel, applying the proposed optimization framework to translational and clinical datasets will make it possible to study how interventions such as neuromodulation or pharmacological treatments alter the underlying dynamical regimes, providing a pathway for using whole-brain models to assess treatment-induced changes in large-scale brain organization.

Complementary to these data-driven extensions, two methodological directions emerge naturally from the evolutionary optimization perspective. First, extending the HICO framework to substantially higher-dimensional parameterizations may further improve biological significance of the solutions. As whole-brain models incorporate finer-grained region and circuits, understanding how curricular structure scales with dimensionality will be essential for promoting stability, generalization, and relevance.

Second, the HICO framework may be combined with distributed attention-based approaches such as BLADE \cite{Shahrzad2023}. Applying BLADE within each optimization phase could selectively reduce the effective search dimensionality while preserving hierarchical structure, thereby compensating for the increased complexity of larger models. Such a hybrid strategy offers a principled way to balance expressivity and efficiency, and makes it possible to compare curriculum-guided attention-based evolution and unconstrained optimization of higher-dimensional models. Exploring this interaction may yield more scalable evolutionary optimization schemes for large, structured dynamical systems.

\section{Conclusion}
\label{sec:Conclusion}

This study demonstrated that the curricular structure of parameter optimization plays a crucial role in generality and relevance of whole-brain biophysical models. 
Biologically motivated structure acts as a powerful inductive bias in the evolutionary optimization: it uncovers parameter regimes that generalize effectively across subjects, and can be used to predict behavior.
Curriculum-informed evolutionary optimization thus provides a principled approach for scaling complex biophysical models. This framework offers a promising direction for future work in large-scale computational neuroscience and other high-dimensional scientific modeling domains.
\vspace*{-2ex}

\section*{Acknowledgments}
This work was supported by National Institutes of Mental Health (NIMH) grant R01MH127608 and a Stanford Maternal and Child Health Research Institute (MCHRI) Faculty Scholar Award to M.S. Funding for the Human Connectome Project data acquisition were provided by the 16 NIH Institutes and Centers that support the NIH Blueprint for Neuroscience Research (as part of the Human Connectome Project, WUMinn Consortium; Principal Investigators: David Van Essen and Kamil Ugurbil; 1U54MH091657) and by the McDonnell Center for Systems Neuroscience at Washington University. 
The authors also gratefully acknowledge Daniel Fink for his engineering contributions to the platform implementation.

\newpage
\bibliographystyle{ACM-Reference-Format}
\bibliography{main}

\newpage
\appendix
\section{Dynamic Mean Field Parameters and Fitness Computation}
\label{app:appendix_dmf_parameters}

This appendix specifies the Dynamic Mean Field (DMF) model parameters, their biological interpretation, default values and ranges, and the procedure used to compute functional connectivity (FC)–based fitness.

\subsection{DMF State Variables and Fixed Constants}
\label{app:dmf_parameters}

Each cortical region $i$ is modeled as a coupled excitatory–inhibitory neural population. The dynamical state is described by synaptic gating variables $S_i(t)$ (excitatory) and $S^I_i(t)$ (inhibitory), evolving according to the DMF equations defined in Section~2.2.

The following constants were fixed across all experiments and adopted from prior DMF studies \cite{Deco2009,Murray2014,Breakspear2017}.

\begin{center}
\begin{tabular}{lll}
\hline
Symbol & Description & Value \\
\hline
$\tau_E$ & Excitatory synaptic time constant & 100 ms \\
$\tau_I$ & Inhibitory synaptic time constant & 10 ms \\
$\gamma$ & Synaptic gain & 0.641 \\
$a_E$ & Excitatory gain slope & 310 nC$^{-1}$ \\
$b_E$ & Excitatory firing threshold & 125 Hz \\
$d_E$ & Excitatory curvature parameter & 0.16 s \\
$I_b$ & Background input current & fixed \\
\hline
\end{tabular}
\end{center}

The sigmoidal transfer function $H(\cdot)$ follows the standard DMF formulation \cite{Deco2009}, mapping total synaptic input to population firing rate.

\subsection{Optimized Parameters per RSN}
\label{app:rsn_params_table}

The Dynamic Mean Field (DMF) model parameters optimized in this study follow the formulation of \citet{Deco2011,Deco2009,Murray2014}.  
Each RSN is associated with an identical ordered block of 20 parameters.  

For clarity, the parameters are grouped into (i) \emph{local circuit parameters} (Table~\ref{tab:local_circuit_params}), governing excitatory–inhibitory dynamics within each region, and (ii) \emph{global and long-range coupling parameters} (Table~\ref{tab:global_coupling_params}), governing noise, synaptic scaling, and interactions between regions.
Reported ranges correspond to bounds used in prior DMF studies; default values refer to canonical settings commonly adopted before optimization.

\begin{table}[t]
\centering
\footnotesize
\begin{tabular}{cllp{3.4cm}}
\hline
\textbf{Index} & \textbf{Parameter} & \textbf{Default / Range} & \textbf{Description} \\
\hline
0  & $a_E$    & $310$ \; [$200$--$400$] & Gain of the excitatory firing-rate input--output function. \\
1  & $b_E$    & $125$ \; [$100$--$150$] & Offset (bias) of the excitatory firing-rate function. \\
2  & $d_E$    & $0.16$ \; [$0.1$--$0.3$] & Curvature parameter controlling slope of excitatory nonlinearity. \\
3  & $W_E$    & $1.0$ \; [$0.5$--$2.0$] & Scaling of recurrent excitatory input. \\
4  & $\tau_E$ & $100$ ms \; [$50$--$200$] & Time constant of excitatory synaptic gating. \\
5  & $a_I$    & $615$ \; [$400$--$800$] & Gain of the inhibitory firing-rate function. \\
6  & $b_I$    & $177$ \; [$150$--$220$] & Offset (bias) of the inhibitory firing-rate function. \\
7  & $d_I$    & $0.087$ \; [$0.05$--$0.15$] & Curvature parameter of inhibitory nonlinearity. \\
8  & $W_I$    & $0.7$ \; [$0.3$--$1.5$] & Scaling of recurrent inhibitory input. \\
9  & $\tau_I$ & $10$ ms \; [$5$--$20$] & Time constant of inhibitory synaptic gating. \\
10 & $w_{EE}$ & $1.4$ \; [$0.5$--$2.5$] & Excitatory-to-excitatory recurrent coupling strength. \\
11 & $w_{EI}$ & $1.0$ \; [$0.5$--$2.0$] & Inhibitory input strength onto excitatory population. \\
12 & $w_{IE}$ & $1.0$ \; [$0.5$--$2.0$] & Excitatory input strength onto inhibitory population. \\
13 & $w_{II}$ & $0.5$ \; [$0.1$--$1.5$] & Inhibitory-to-inhibitory recurrent coupling strength. \\
\hline
\end{tabular}
\caption{Local circuit parameters optimized per RSN. These parameters control the excitatory--inhibitory population dynamics within each cortical region.}
\label{tab:local_circuit_params}
\end{table}

\begin{table}[t]
\centering
\footnotesize
\begin{tabular}{cllp{3.4cm}}
\hline
\textbf{Index} & \textbf{Parameter} & \textbf{Default / Range} & \textbf{Description} \\
\hline
14 & $I_b$      & $0.382$ \; [$0.2$--$0.6$] & Background (external) input current to excitatory population. \\
15 & $J$        & $0.15$ \; [$0.05$--$0.3$] & Synaptic scaling factor converting firing rate to synaptic current. \\
16 & $\gamma$   & $0.641$ \; [$0.3$--$1.0$] & Gain of excitatory synaptic gating dynamics. \\
17 & $\gamma_I$ & $1.0$ \; [$0.5$--$1.5$] & Gain of inhibitory synaptic gating dynamics. \\
18 & $\sigma$   & $0.01$ \; [$0.001$--$0.05$] & Noise amplitude driving stochastic synaptic fluctuations. \\
19 & $g$        & $2.5$ \; [$0$--$5$] & Global coupling strength applied to structural connectivity. \\
\hline
\end{tabular}
\caption{Global and long-range coupling parameters optimized per RSN. These parameters regulate noise, synaptic scaling, and interregional communication through the structural connectome.}
\label{tab:global_coupling_params}
\end{table}

\paragraph{Notes on parameter usage.}
In the heterogeneous formulation, Tables~\ref{tab:local_circuit_params} and \ref{tab:global_coupling_params} together define a 20-dimensional parameter block that is replicated independently across the seven RSNs, yielding a total of 140 free parameters.  
In the homogeneous baseline, a single shared instance of this block is optimized and applied uniformly across all RSNs.  
All parameters were represented internally in normalized coordinates in $[0,1]$ and affinely mapped to the ranges listed above during fitness evaluation.

\subsection{Simulation Pipeline}
\label{app:pipeline}

For a candidate parameter vector $\Theta$, the DMF system was simulated for five minutes of biological time using Euler–Maruyama integration \cite{Higham2001} with step size $\Delta t = 0.1$ ms. The resulting excitatory synaptic activity $S_i(t)$ was transformed into BOLD signals using the Balloon–Windkessel hemodynamic model \cite{Friston2003}.

Functional connectivity matrices $\widehat{\mathrm{FC}}(\Theta)$ were computed as Pearson correlations between regional BOLD time series, matching the construction of empirical FC from resting-state fMRI.

\subsection{Fitness Definition and Evaluation}
\label{app:fitness}

Although the loss function in Eq.~(2) is written as a minimization objective, evolutionary optimization was performed by maximizing the Pearson correlation between empirical and simulated FC matrices:
\[
\mathrm{fitness}(\Theta) = \mathrm{corr}\!\left(\mathrm{vec}(\mathrm{FC}),\,\mathrm{vec}\!\left(\widehat{\mathrm{FC}}(\Theta)\right)\right).
\]
Throughout the paper, the terms \emph{fitness} and \emph{score} refer to this correlation value.

For leave-one-out (LOO) evaluation, subject-specific parameter vectors were aggregated using a two-sided trimmed mean ($p=0.1$), producing a cohort-level parameter estimate $\tilde{\theta}^{(-j)}$ that was then evaluated on the held-out subject using the same FC-based fitness computation. The moments-based fitness implementation follows that of \citet{Murray2014}.

\subsection{Methodological Clarifications}
\label{app:clarifications}

All simulations used subject-specific structural connectivity matrices derived from diffusion-weighted imaging. Behavioral prediction analyses and permutation-based significance testing were performed after all subject-level evolutionary runs were completed and did not influence optimization dynamics. This separation ensures that the learned DMF parameters define task-agnostic whole-brain dynamical models suitable for downstream analysis across multiple behavioral domains.

\end{document}